\pdfoutput=1
\documentclass[11pt]{article}

\usepackage[preprint]{acl}

\usepackage{times}
\usepackage{latexsym}

\usepackage[T1]{fontenc}
\usepackage[utf8]{inputenc}
\usepackage{microtype}
\usepackage{inconsolata}
\usepackage{graphicx}
\usepackage{tikz}
\usepackage{pdflscape}
\usepackage{booktabs}
\usepackage{multirow}
\usepackage{amsmath}
\usepackage{float}

\usetikzlibrary{positioning}

\title{Revisiting Funnel Transformers for Modern LLM Architectures with Comprehensive Ablations in Training and Inference Configurations}

\author{
  \textbf{DongHyun Choi\textsuperscript{1}},
  \textbf{Lucas Spangher\textsuperscript{1,2}},
  \textbf{Chris Hidey\textsuperscript{1}},
  \\ % Manual line break (adjust layout as needed)
  \textbf{Peter Grabowski\textsuperscript{1,3}},
  \textbf{Ramy Eskander\textsuperscript{1}}
  \\ % End of author list
  \\ % Blank line separator
  \textsuperscript{1}Google,\ 
  \textsuperscript{2}Massachusetts Institute of Technology,\ 
  \textsuperscript{3}University of California, Berkeley
  \\ % End of affiliation line
  \small{
    \textbf{Correspondence:} \href{mailto:dhchoi@google.com}{dhchoi@google.com}, \href{mailto:reskander@google.com}{reskander@google.com}
  }
}

\begin{document}
\maketitle

\begin{abstract}
Transformer-based Large Language Models, which suffer from high computational costs, advance so quickly that techniques proposed to streamline earlier iterations are not guaranteed to benefit more modern models.  Building upon the Funnel Transformer proposed by Dai and Le (2020), which progressively compresses intermediate representations, we investigate the impact of funneling in contemporary Gemma2 Transformer architectures. We systematically evaluate various funnel configurations and recovery methods, comparing: (1) standard pretraining to funnel-aware pretraining strategies, (2) the impact of funnel-aware fine-tuning, and (3) the type of sequence recovery operation. Our results demonstrate that funneling creates information bottlenecks that propagate through deeper network layers, particularly in larger models (e.g., Gemma 7B), leading to at times unmanageable performance lost. However, carefully selecting the funneling layer and employing effective recovery strategies, can substantially mitigate performance losses, achieving up to a 44\% reduction in latency. Our findings highlight key trade-offs between computational efficiency and model accuracy, providing practical guidance for deploying funnel-based approaches in large-scale natural language applications.

\end{abstract}

\section{Introduction}

In recent years, Transformer architectures have revolutionized natural language processing (NLP) by enabling models to capture complex dependencies within sequences. However, this capability comes at a significant computational cost, particularly when processing lengthy sequences, as the self-attention mechanism scales quadratically with sequence length. This scaling issue poses challenges for deploying large language models (LLMs) at production scales in real-world applications.

 Building on the foundational work by Zihang Dai and Quoc Le (2020), who proposed the Funnel Transformer to progressively reduce sequence length by pooling intermediate representations, we extend their approach to modern transformer architectures. Dai and Le demonstrated their concept using the BERT architecture, which, despite its groundbreaking impact, has since become antiquated compared to current state-of-the-art transformer-based models.

In this work, we extend the Funnel Transformer concept to contemporary architectures by specifically testing modern Gemma2 models. Unlike BERT, the Gemma2 family includes Mixture-of-Experts (MoE), of which it is far from given that they should respond similarly to the impact of funneling.

We further distinguish our study through a detailed experimental design that systematically evaluates key aspects of funneling. Specifically, we perform rigorous ablations to determine: 

\begin{enumerate}
    \item At which layer to pool intermediate representations (while targeting a maximum acceptable performance degradation of approximately 5\%) 
    \item At which layer it is optimal to recover the full sequence length. 
    \item We investigate three training scenarios: pretraining transformers with funneling integrated, fine-tuning pretrained transformers exclusively with funneling, and performing inference with funnel transformers without prior pretraining or fine-tuning.
\end{enumerate}

Through these comprehensive evaluations, we aim to quantify the performance trade-offs introduced by funneling and clearly delineate the circumstances under which funneling yields significant computational benefits while maintaining acceptable performance degradation. Our findings provide guidance on effectively leveraging funneling to balance efficiency gains with minimal performance sacrifices in practical, large-scale LLM deployments.

\section{Background Literature}

The success of Transformer-based large language models (LLMs) has come with steep computational costs, motivating research into more efficient architectures \citep{tay2022efficient}. A prominent strategy is the Mixture-of-Experts (MoE) paradigm, which expands model capacity by partitioning parameters into multiple “experts” and activating only a few per input token \citep{zuo2022taming} MoE Transformers \cite{shazeer2017outrageously} (e.g. Switch Transformer) have achieved trillions of parameters with manageable computation per token \citep{fedus2022switch}. However, scaling up MoE models introduces new challenges: the routing mechanism can cause overhead, and the sheer model size complicates deployment \citep{lu2024notallexperts}. 

To address these issues, recent work explores funneling architectures, encoder-only Transformer variants, and other innovations to boost computational efficiency in MoE LLMs \citep{du2022glam}. These methods aim to reduce inference/training cost (FLOPs), memory footprint, or energy consumption, while preserving the accuracy gains of large models \citep{jin2024efficient}. 

While architectural innovations like MoE increase parameter efficiency, another option is choosing the right transformer architecture for the task. In particular, encoder-only models (akin to BERT \citep{devlin2019bert}) can be more efficient than the typical decoder-style LLMs in certain scenarios. Encoder-only Transformers process the entire input sequence bidirectionally and output either a sequence of representations or a single pooled representation. 

The Funnel-Transformer \citep{dai2020funnel} uses a down-sampling encoder that gradually compresses the token sequence to shorter hidden representations, saving computation at higher layers. Intuitively, not all intermediate token positions are needed for high-level understanding, especially for tasks like classification that ultimately require a single output. By “funneling” the sequence, deeper layers operate on a smaller set of hidden states, which substantially cuts down self-attention and FFN costs in those layers. The compressed representation is then optionally up-sampled by a lightweight decoder if token-level output is required (e.g. for a language modeling objective), much like an auto-encoder. Since their inception, they have been used in vision \citep{heo2021rethinking}, pose-estimation \citep{li2022exploiting}, and recommender systems \citep{liu2023linrec}.

Funnel models may be better suited for hybrid architectures that incorporate retrieval or external knowledge, like FunnelRAG \cite{zhao2025funnelragcoarsetofineprogressiveretrieval}, RaSeRec \citep{zhao2024raserec}, or Condenser \citep{gao2021condenser}. In a deployment scenario, one might use a retriever to fetch relevant documents and then an LLM to process them. A funnel Transformer could take those documents (maybe 100 pages total) and compress them into a single vector that a decoder then uses to generate an answer. This two-stage funnel+decoder system would likely be faster and more memory-efficient than a giant end-to-end model that attends to every word of every page throughout. 

As we have seen, a recurring theme in these efficiency methods is sparsity – not all parts of a model or input need equal processing. 
\section{Model Design}

\begin{figure*}
    \centering
    \includegraphics[width=.9\textwidth]{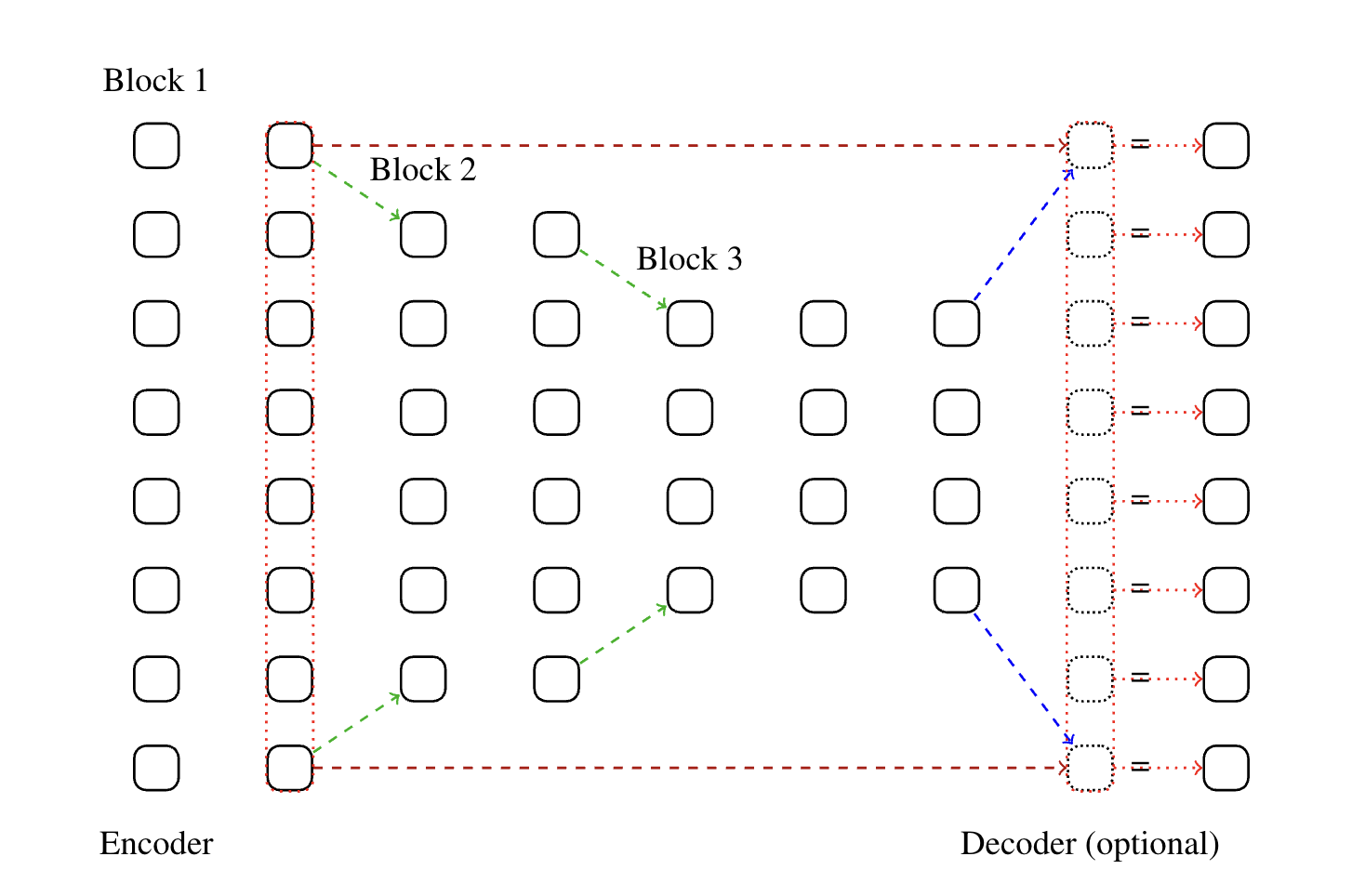}
    \caption{A representation of a funnel architecture with seven encoder layers and two funneling operations -- one at the ``last'' pre-funnling layer, layer 3, and one at the recovery sequence layer, layer 7. (For the purposes of this caption, layers are 1-indexed.)}
    \label{fig:funnel_model}
\end{figure*}

In our experiments, we perform all training on Gemma2b and 7b models. We conduct a limited grid search and produce a hyperparameter configuration that we kept constant across model architectures, listed in the Appendix. Please find the hyperparameters for Gemma2 2B at Table \ref{table:hyperparameters} and the hyperparamaters for Gemma2 7B at Table \ref{tab:hyperparameters_7b}.  
\section{Experimental Setup}

\subsection{Benchmarks studied}

In this work, we primarily report results on the \textbf{General Language Understanding Evaluation (GLUE)} benchmark and the \textbf{CoNLL-2003 Named Entity Recognition (NER)}. \textbf{GLUE} is a collection of nine natural language understanding tasks designed to evaluate and analyze the performance of models across diverse linguistic phenomena. Introduced by Alex Wang et al. in 2018, GLUE includes tasks such as single-sentence classification, similarity and paraphrase detection, and natural language inference, providing a comprehensive platform for assessing general language understanding capabilities. We include GLUE as a \textit{sentence level classification} task.

The \textbf{CoNLL-2003 Named Entity Recognition (NER)} \citep{sang2003introductionconll2003sharedtask} benchmark is a widely utilized dataset for evaluating NER systems in English. It consists of text annotated with four entity types: persons, locations, organizations, and miscellaneous entities. Models are assessed using the F1 score, which balances precision and recall. We present the average scores between the english and german NER. We include NER as a \textit{token-level classification} task.

Thirdly, we report results on an internal industry benchmark called \textbf{WebAnswers Classification}. In order to maintain confidentiality and anonymity, we opt not to describe the benchmark in this reviewing phase, only to say that it is a carefully curated benchmark relevant to the goals of our company. The metric that we report from this benchmark is the ROC AUC of the classification task. WebAnswers, like GLUE, is \textit{sentence-level} classification.

\subsection{Testing the effect of pretraining on funnel transformer setup}

We compare whether pretraining the transformer in a funnel-aware setup positively impacts the scores. In all cases, we pretrain the Gemma2 models for 100k steps on an internal benchmark; in the funnel-aware pretraining setup, we perform a max-pool two-token funnel once at layer 2 and then propagate the continue with the reduced dimension through the rest of the transformer. Once pretrained, we test the results of applying a max-pool two-token funnel to every even layer in a 16 layer Gemma2 model: that is, layers 2, 4, 6, 8, ..., 16. Please note that when the max-pool funnel is applied to the second layer, it matches the funnel-aware pretraining configuration exactly. We report results on sentence level tasks only only, and thus we do not recover the sequence length in any of the experiments. 

\subsection{Fine-tuning a funnel-aware setup}

We fine-tune the funnel-aware Gemma2 2b and 7b models on all GLUE benchmarks and present an average. We fine-tune for 2000, 4000, and 6000 steps and perform a two-token max-pool funnel at every even layer of a 16 layer 2B Gemma2 model, and similarly at every even layer of the 28 layer 7B Gemma2 model. Here, we profile latency on the Gemma2 2B models by measuring wall-clock completion across the benchmark. Similarly to the pretraining experiments, we report results on sentence level classification tasks only, and also hold constant the lack of sequence length recovery. 

\subsection{Optimizing the type of recovery operation}

We ablate among different types of operations to recover the full sequence length \textit{at the last layer} of the Gemma2 2B model. In all variants, we tile the intermediate activations of the last funnel layer to recover the full dimension (that is, if the layer is half the original dimension due to funneling, we repeat each activation twice to create a layer the original dimension; $(1, 3, 4) \rightarrow (1, 1, 3, 3, 4, 4)$. We test the following variants: 

\begin{enumerate}
    \item \textbf{ Sum with first layer (baseline)}: similar to the operation used in \citep{dai2020funnel}, we add the tiled funnel activation to the output activations of the transformer's first layer. 
    \item \textbf{Sum with last layer}: we add the tiled funnel activation to the output activations of the final full layer before funneling was performed.
    \item \textbf{Sum with all the previous layers' max}: we add the tiled funnel activation to the maximum index-wise activation of all prior layers. 
    \item \textbf{Sum with all the previous layers' avg}: we add the tiled funnel activation to the average of the activations of all prior layers.
    \item \textbf{Average with last layer}: we average the tiled funnel activation with the activation of the final pre-funnelled layer. 
    \item \textbf{Max(Last, Now)}: we take the index wise maximum of the last pre-funnelled layer and the tiled funnel activation. 

\end{enumerate}
 
We compute performance across even steps of a two-token max-pool funnel that is applied throughout the 16 layers of the architecture. We present scores on the NER benchmark, as it is a token level task and thus relevant to recover the full sequence length. 

% \subsection{Optimizing the type of pooling operation}

% Due to limited compute and the desire to test more pooling operations, we perform a random search on the funnelled layer of a Gemma2 model and the pooling operation performed, and specific GLUE task. We perform a linear regression controlling for the funneling layer, and present the regression coefficient weights. Those activations that are significantly positive, we list as successful pooling operations. We test: 

% \begin{enumerate}
%     \item first token pool
%     \item max pool
%     \item attention pool
%     \item avg. (max pool, attention pool)
%     \item avg. (max pool, first token pool)
%     \item avg. (attention pool, first token pool)
%     \item avg. (max pool, attention pool, first token pool)
%     \item \( l_{p=2} \) norm: (max pool, attention pool, first token pool)
%     \item \( l_{p=2} \) norm: (max pool, attention pool)
%     \item \( l_{p=2} \) norm: (max pool, first token pool)
%     \item stochastic pool
%     \item avg. (max pool, stochastic pool)
%     \item \( l_{p=2} \) norm: (max pool, attention pool, gated pool)
%     \item gated pool
%     \item avg. (stochastic pool, gated pool)
% \end{enumerate}

\section{Results}

\subsection{Effect of pretraining}

\begin{figure*}[htb]
    \centering
    \includegraphics[width=\textwidth]{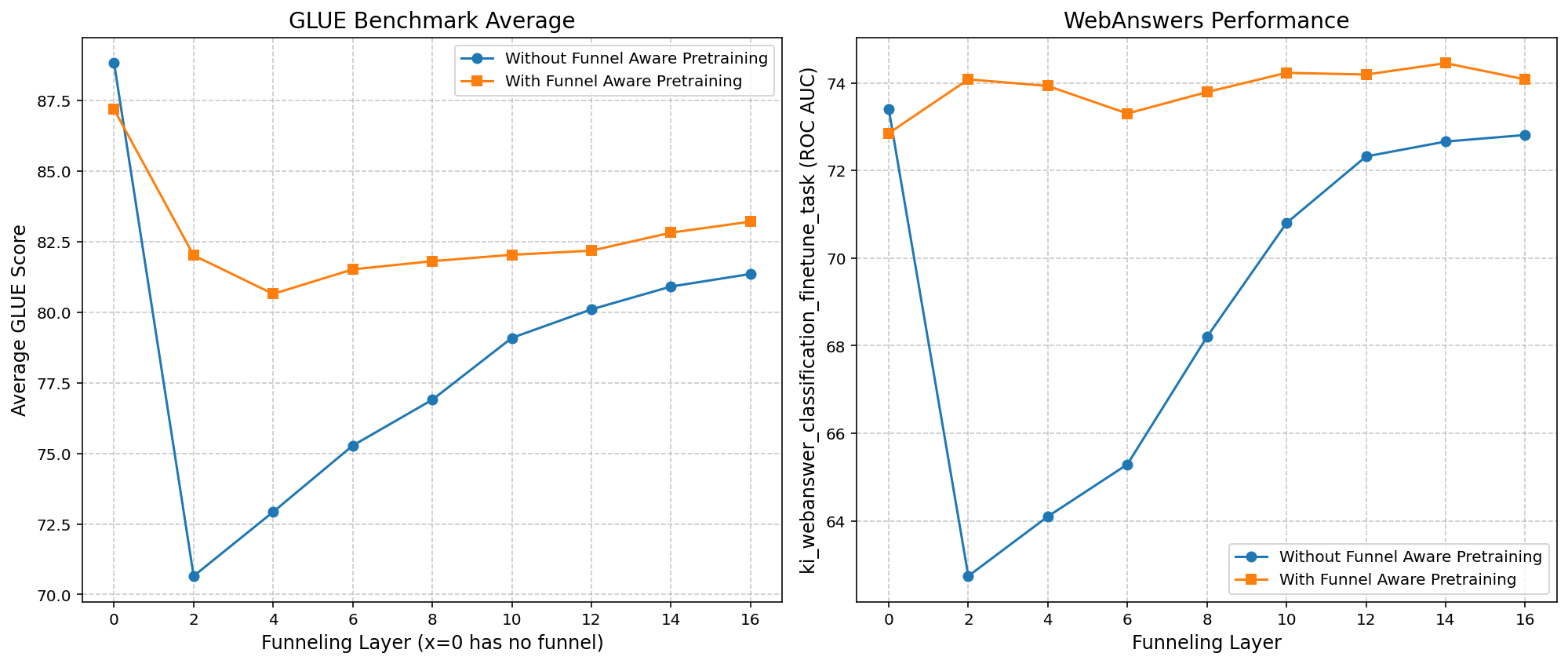}
    \caption{Performance on (a) GLUE benchmark (Average GLUE Score) and (b) WebAnswers ROC AUC as a function of the funnel recovery layer. The x=0 point corresponds to the model without funneling. Solid lines represent models trained with normal pretraining (``Without Funnel Aware Pretraining'') and funnel-aware pretraining (``With Funnel Aware Pretraining'').}
    \label{fig:pretraining}
\end{figure*}

Figure~\ref{fig:pretraining} displays two performance metrics as a function of the funnel recovery layer. The left panel shows the average GLUE score, while the right panel shows ROC AUC on the WebAnswers dataset. In each plot, the x-axis includes a 0 point that represents the “No Funnel” case. For the GLUE benchmark, the normal pretraining curve has a value of 88.81 at x=0, whereas the funnel-aware pretraining curve starts at 87.17. Similarly, in the WebAnswers plot the “No Funnel” baseline is 73.40 for normal pretraining and 72.85 for funnel-aware pretraining. For both metrics, performance is tracked over increasing funnel recovery layers (2, 4, \dots, 16).

\begin{table*}[h!]
\centering
\caption{Performance metrics for different funnel configurations. The left-most column, "Funnel Layer", corresponds with the x-axis of our charts and indicates at which layer the two-token max-pool funnel is applied.}
\label{tab:pretraining_performance}
\resizebox{\linewidth}{!}{%
\begin{tabular}{@{}lccccccccccc@{}}
\toprule
\textbf{\begin{tabular}[c]{@{}c@{}}Funnel\\ Layer\end{tabular}} & \textbf{\begin{tabular}[c]{@{}c@{}}stsb\\ (Spearman)\end{tabular}} & \textbf{\begin{tabular}[c]{@{}c@{}}cola\\ (Acc.)\end{tabular}} & \textbf{\begin{tabular}[c]{@{}c@{}}qqp\\ (Acc.)\end{tabular}} & \textbf{\begin{tabular}[c]{@{}c@{}}qnli\\ (Acc.)\end{tabular}} & \textbf{\begin{tabular}[c]{@{}c@{}}sst2\\ (Acc.)\end{tabular}} & \textbf{\begin{tabular}[c]{@{}c@{}}rte\\ (Acc.)\end{tabular}} & \textbf{\begin{tabular}[c]{@{}c@{}}mrpc\\ (Acc.)\end{tabular}} & \textbf{\begin{tabular}[c]{@{}c@{}}mnli\\ (m-Acc.)\end{tabular}} & \textbf{\begin{tabular}[c]{@{}c@{}}mnli\\ (mm-Acc.)\end{tabular}} & \textbf{\begin{tabular}[c]{@{}c@{}}webanswer \\ classification\\ (ROC AUC)\end{tabular}} & \textbf{\begin{tabular}[c]{@{}c@{}}GLUE \\ Avg.\end{tabular}} \\ 
\midrule
\multicolumn{12}{c}{\textbf{normal pretraining}} \\
\midrule
2  & 59.79 & 66.73 & 83.89 & 78.56 & 84.98 & 53.43 & 70.83 & 68.65 & 69.03 & 62.74 & \textbf{70.65} \\
4  & 67.82 & 67.21 & 84.00 & 80.62 & 86.70 & 55.60 & 66.42 & 73.08 & 74.89 & 64.10 & \textbf{72.93} \\
6  & 77.47 & 67.11 & 85.11 & 82.94 & 85.67 & 55.96 & 72.30 & 75.13 & 75.77 & 65.29 & \textbf{75.27} \\
8  & 84.51 & 65.77 & 86.79 & 84.02 & 88.07 & 55.60 & 72.79 & 76.72 & 77.78 & 68.20 & \textbf{76.89} \\
10 & 87.02 & 69.61 & 86.90 & 86.34 & 88.88 & 56.68 & 78.19 & 78.95 & 79.19 & 70.80 & \textbf{79.08} \\
12 & 86.95 & 75.55 & 86.84 & 86.40 & 89.56 & 59.57 & 78.43 & 78.66 & 78.89 & 72.32 & \textbf{80.09} \\
14 & 87.64 & 76.99 & 86.54 & 86.75 & 89.33 & 59.57 & 81.37 & 79.74 & 80.19 & 72.66 & \textbf{80.90} \\
16 & 87.83 & 76.70 & 86.53 & 87.92 & 89.56 & 59.57 & 82.35 & 80.66 & 81.01 & 72.81 & \textbf{81.35} \\
\midrule
\multicolumn{12}{c}{\textbf{funnel aware pretraining}} \\
\midrule
2  & 88.00 & 76.80 & 88.17 & 88.28 & 90.25 & 59.57 & 85.78 & 80.22 & 80.98 & 74.08 & \textbf{82.01} \\
4  & 86.91 & 74.59 & 87.59 & 87.50 & 90.37 & 59.93 & 79.17 & 79.60 & 80.12 & 73.93 & \textbf{80.64} \\
6  & 87.11 & 75.26 & 87.37 & 87.31 & 90.37 & 60.65 & 84.56 & 80.18 & 80.78 & 73.30 & \textbf{81.51} \\
8  & 87.67 & 76.51 & 87.29 & 88.39 & 90.83 & 58.48 & 86.52 & 79.77 & 80.75 & 73.79 & \textbf{81.80} \\
10 & 88.78 & 75.65 & 87.75 & 87.81 & 89.79 & 61.37 & 85.54 & 80.44 & 81.11 & 74.23 & \textbf{82.03} \\
12 & 88.40 & 76.03 & 87.35 & 88.12 & 90.94 & 61.37 & 85.29 & 80.55 & 81.50 & 74.19 & \textbf{82.17} \\
14 & 88.55 & 76.03 & 87.73 & 87.86 & 90.83 & 67.51 & 84.80 & 80.70 & 81.28 & 74.45 & \textbf{82.81} \\
16 & 88.35 & 78.04 & 87.99 & 88.45 & 90.83 & 66.43 & 87.01 & 80.55 & 81.15 & 74.08 & \textbf{83.20} \\
\bottomrule
\end{tabular}%
}
\end{table*}

% \end{landscape}

\subsection{Fine tuning a funnel-aware setup}

\begin{figure*}[htb]
    \centering
    \includegraphics[width=\textwidth]{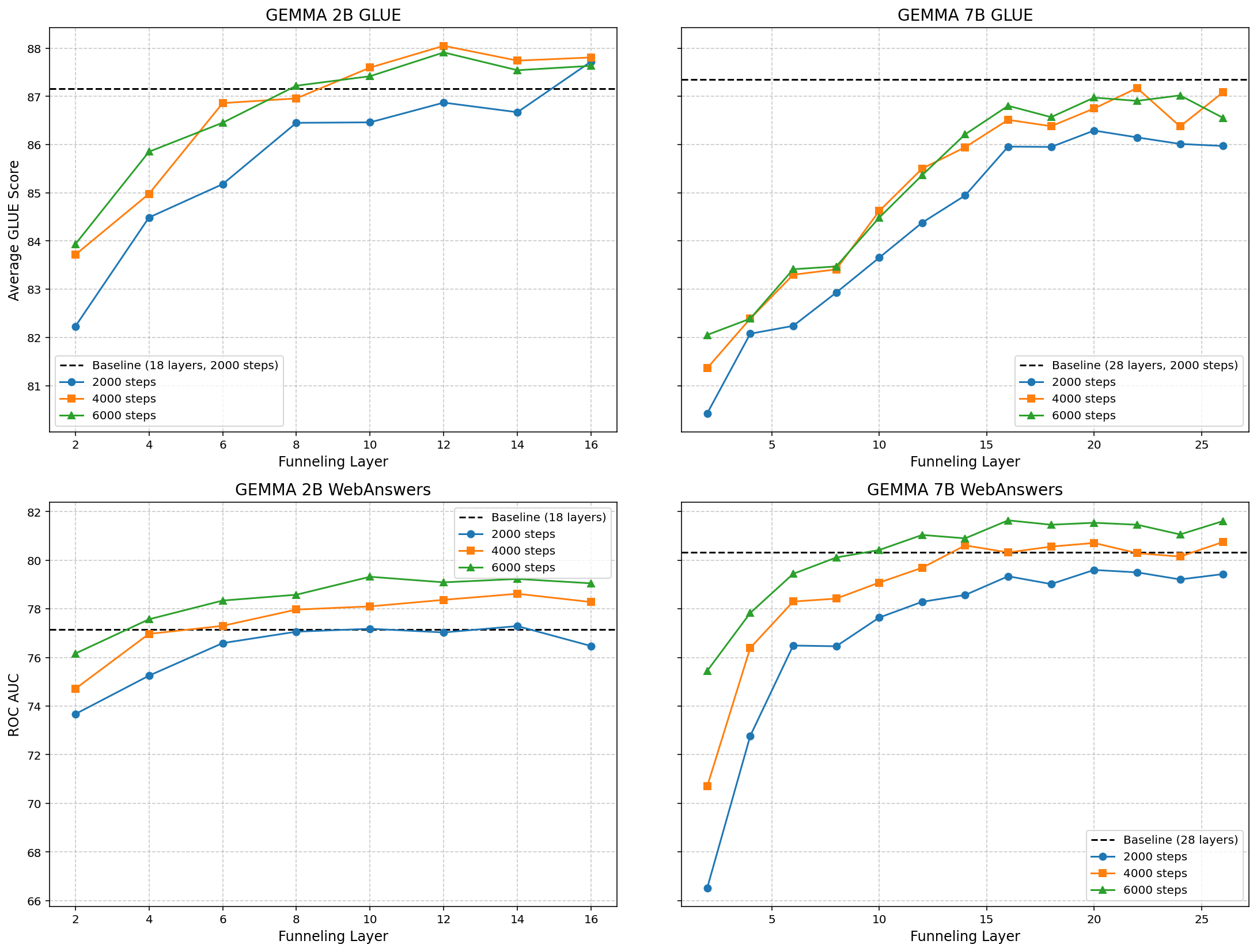}
    \caption{Performance of Gemma2 2B (left) and Gemma2 7B (right) on GLUE (top row) and WebAnswers ROC AUC (bottom row) are plotted against successive layers at which 2-token funnel is applied within each architecture.  Solid lines correspond to models performance with different numbers of fine tuning aware steps, whereas dotted lines correspond to baselines in which no funneling is applied.}
    \label{fig:finetuning}
\end{figure*}

Please see figure \ref{fig:finetuning} for a depiction of the effects of fine-tuning funnel aware architectures across Gemma 2b and 7b models. More fine-tuning seems to help models perform better. 

Figure~\ref{fig:finetuning} presents a set of four subplots comparing the performance of Gemma2 2B and Gemma2 7B models on two tasks: the GLUE benchmark (top row) and the WebAnswers ROC AUC (bottom row). In each subplot, the x-axis denotes the funnel recovery layer, with an x=0 point corresponding to the “No Funnel” case. Two performance curves are shown in each panel: one for models trained without funnel-aware pretraining (normal pretraining) and one with funnel-aware pretraining. 

In the GLUE subplots (top row), the normal pretraining curve starts at an 88.81 score for Gemma2 2B and 87.17 for Gemma2 7B at x=0, while the funnel-aware pretraining curves start at 87.17 and lower, respectively. For the WebAnswers subplots (bottom row), the “No Funnel” baseline is 73.40 for normal pretraining and 72.85 for funnel-aware pretraining, with performance measured as WebAnswers ROC AUC. Recovery data is plotted for increasing funnel recovery layers (2, 4, 6, \dots, 16 for Gemma2 2B and 2, 4, \dots, 26 for Gemma2 7B).

While the baseline of Gemma7b is higher than the baseline of Gemma2b, the performance of Gemma7b seems to suffer more overall than the performance of Gemma2b.

\begin{figure}[htb]
    \centering
    \includegraphics[width=\columnwidth]{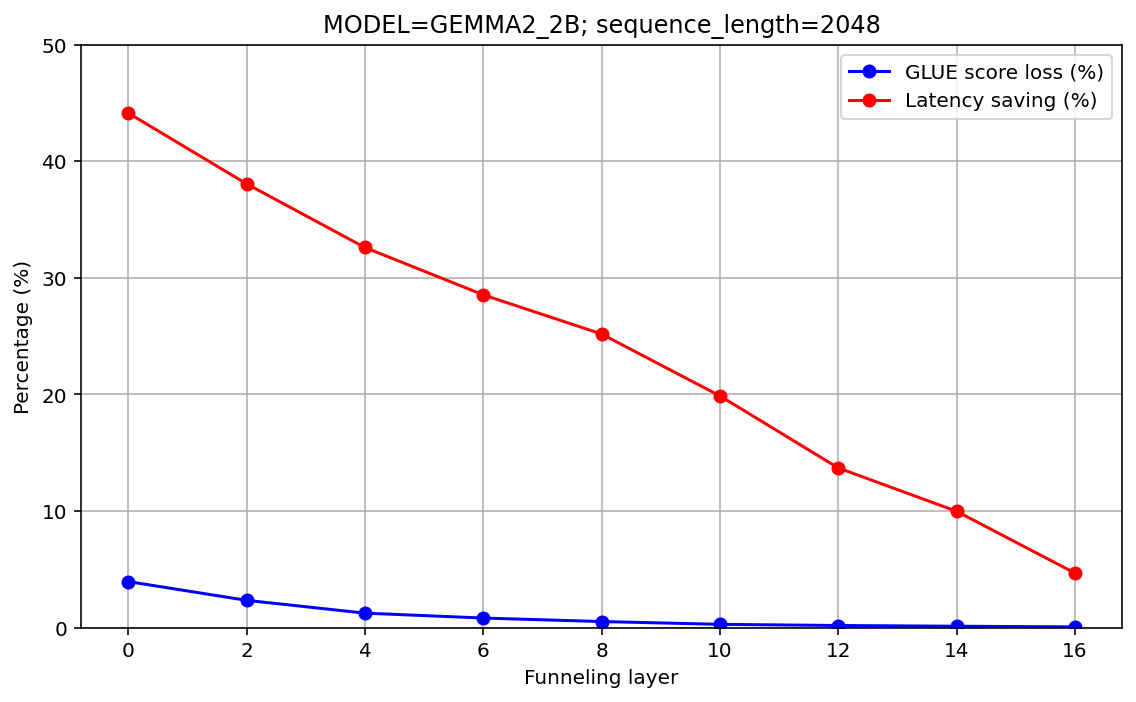}
    \caption{Comparison of latency versus performance gains.}
    \label{fig:latency}
\end{figure}

As shown in our latency results, i.e. Figure \ref{fig:latency} plot, introducing the funneling layer at earlier stages provides substantial latency savings, with a peak of over 40\% when funneling is applied at or near layer 0. However, as the funneling layer increases (moving further into the model), the latency gains steadily diminish. By layer 16, the latency savings drop to around 5\%. This trend suggests that while early funneling can significantly speed up inference, its benefits taper off if the funneling is applied deeper in the network.

\subsection{Optimizing the type of recovery operation}

\begin{figure*}[htb!]
    \centering
    \includegraphics[width=.75\textwidth]{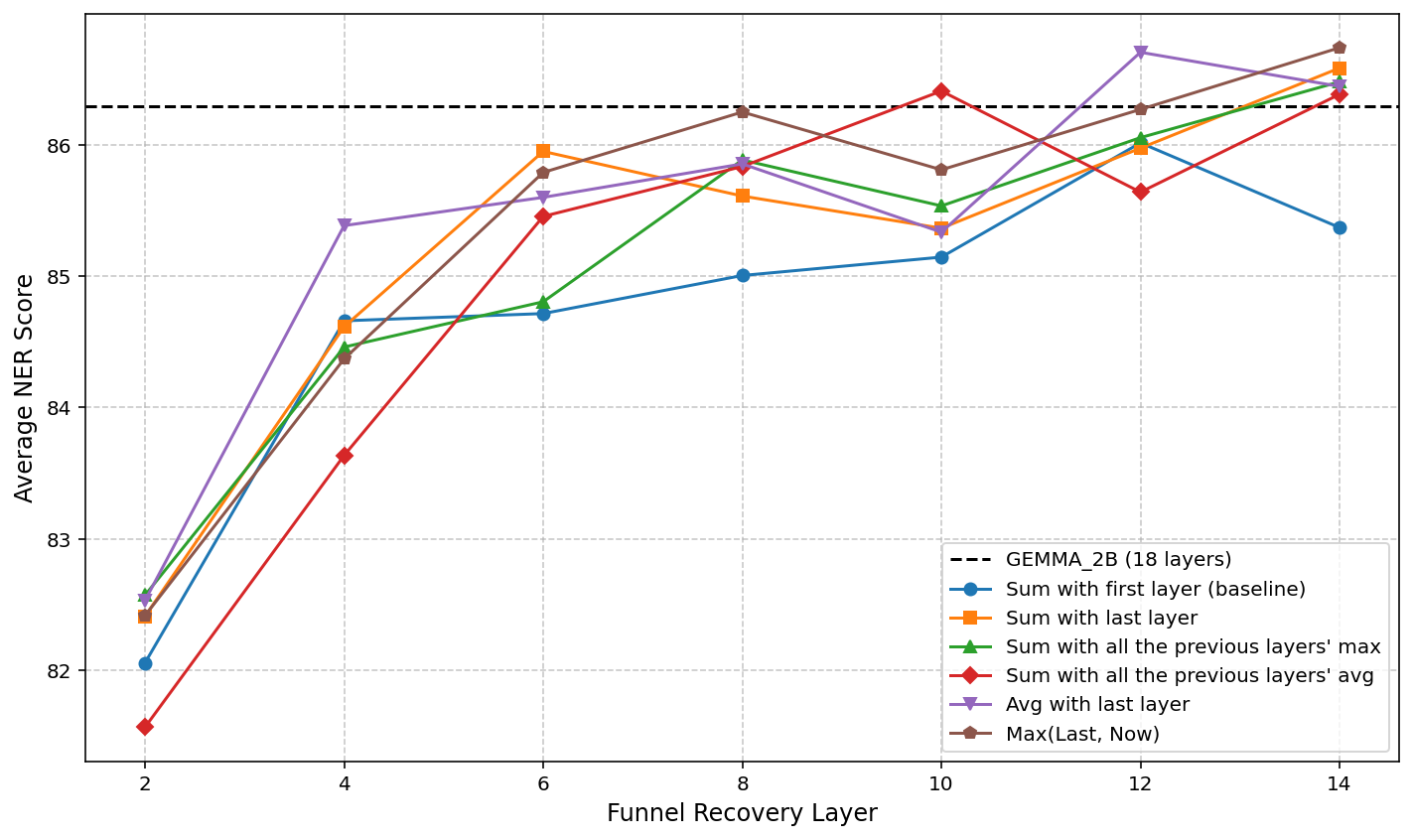}
    \caption{The effect of different recovery operations on NER performance.}
    \label{fig:recover_layer}
\end{figure*}

Please see Figure \ref{fig:recover_layer} for a depiction of the effects of different recover operations on a Gemma 2b model. 

% \subsection{Regression results for pooling operation}
\section{Discussion}

\subsection{Impact of Pretraining on Accuracy and Quality Drop}

Our results indicate that the model's accuracy follows a V-curve, where performance initially drops as the funnel configuration diverges from the original pretraining configuration. One plausible explanation is that pretraining helps to blunt this drop in quality. Specifically, the pretraining procedure incorporates a second layer that is funnel-aware, which appears to counterbalance the information loss introduced by the funnel configuration at that particular layer.

Interestingly, when no funneling is applied (x=0), models trained with normal pretraining outperform those with funnel-aware pretraining, causing the corresponding performance curves to cross over. This suggests that in the absence of any funnel-induced modifications, the additional complexity introduced by funnel-aware pretraining does not confer a benefit and may even be detrimental.

The shapes of the performance curves in both the GLUE and WebAnswers plots exhibit the characteristic V-pattern, with an initial decline in performance followed by a recovery as the funnel configuration is further adjusted. This recovery might be indicative of the funnel mechanism’s potential to mitigate overfitting in the full models. In fact, the observed improvement in performance at higher funnel recovery layers could be a consequence of reduced overfitting, a phenomenon that has been reported in the literature in contexts such as model quantization \citep{biderman2024loralearnsforgets}, where modifying the model architecture can sometimes lead to an increase in performance.

\subsection{Information Bottlenecks}

Peppered throughout our results are the effect of information bottlenecks: that is, the extend to which information is restricted and its impact on performance.

First, \textbf{larger models are impacted more by funneling.} That is, funneling exhibits a notably more pronounced negative effect on the larger and more complex Gemma2 7B model compared to Gemma2 2B. Due to its increased width (number of neurons per layer) and deeper architecture (10 additional layers), the same funnel operation causes greater information restriction when applied to Gemma2 7B. As an example, funneling at the same recovery layer — for instance, layer 6 — aggregates significantly more activations in Gemma2 7B due to its wider layers, and the compressed information propagates across more layers (28 versus 18 in Gemma2 2B). Consequently, this results in a more substantial performance degradation in Gemma2 7B under the same stated funneling operations. This observation indicates that while funneling can offer notable computational speedups by reducing the dimensionality and number of processed activations, careful calibration is required. In particular, the benefits of speedup must be weighed against the degree of performance loss, which can be substantially greater for larger and more complex models.

Additionally, we observed a clear trend that\textbf{ performance consistently degrades as funneling is introduced at earlier layers}. This phenomenon arises because restricting information early in the network negatively affects the quality of learned representations throughout all subsequent layers. Thus, the timing of the funnel operation significantly influences performance outcomes, underscoring the necessity of strategically selecting later funnel recovery layers if maintaining task accuracy is a priority.

Moreover, among the recovery strategies evaluated, \textbf{averaging the unfunneled layer's output with the last layer emerged as the most stable and effective approach}. This averaging operation effectively provides better information passthrough, as it combines the detailed, uncompressed activations from earlier layers with the highly abstract representations in later layers. Compared to other methods such as direct concatenation or max-pooling, averaging ensures a more balanced preservation of both fine-grained details and abstract patterns, which likely explains its superior stability and performance.

exhibits more information compression due to its larger scale. This compression likely contributes to the observed performance degradation. Moreover, there is an indication that Gemma2 2B might be overfitting, as some scenarios reveal that modifications such as quantization or other adjustments can lead to the tuned model outperforming its original full model counterpart.

\subsection{Hyperparameter Tuning and Model Scale Effects}
One more prosaic cause of the performance differences between the Gemma2 2B and Gemma2 7B models may be a result of differences in hyperparameter tuning. We had enough compute to thoroughly hyperparamater tune the Gemma2 2B model, whereas we could only explore locally around default recommendations for the Gemma2 7B. Therefore, the Gemma2 7B model, despite recovering the performance of the full model, saw its performance overall surpassed by the 2B model. We argue that the results are still valid within each model, as both models are compared apples to apples. 

\subsection{Variability Across Random Seeds}
It is important to note that our reported GLUE scores are averaged over 8 to 9 runs with different random seeds. This averaging process is crucial for reducing the variability in performance outcomes and ensuring that the results are robust. While additional plots showing the variance across different seeds could further illustrate these effects, the averaged scores already provide a reliable measure of performance consistency.

\subsection{Limitations}

Our study has several limitations worth noting. Firstly, our experiments \textbf{exclusively used the Gemma2 model family}, limiting the generalizability of our conclusions to other model architectures. Additionally, we \textbf{only employed pretraining that was aware of a single funneling layer;} the effects of incorporating additional funnel-aware pretraining layers remain unexplored. We \textbf{did not explore more aggressive funneling configurations}, such as 4-step funneling or multiple funneling layers in one architecture. Furthermore, our analysis \textbf{did not consider architectures utilizing mixture-of-experts}, which could significantly influence how information compression impacts performance. We also \textbf{restricted our study to a single pooling operation,} leaving alternative pooling mechanisms unexamined. Finally, we \textbf{did not directly compare our funneling approach to other pruning or information bottleneck methods,} limiting insights into its relative efficacy in broader contexts.

\section{Conclusion}

Large Language Models (LLMs) present significant computational demands, necessitating ongoing optimization efforts. This study revisited the Funnel Transformer architecture, investigating its application to the contemporary Gemma model family. Experimentation with varied funnel configurations, under both standard and funnel-aware pretraining, on benchmarks like GLUE and WebAnswers, revealed that aggressive funneling creates information bottlenecks, which can degrade performance, particularly in larger models. However, strategic funnel placement and output averaging of compressed and uncompressed layers effectively mitigated these losses. Averaging proved more robust than other recovery methods, likely by balancing detailed and abstract feature integration. Our results underscore the trade-off between computational efficiency and performance in funneling. Future work should explore enhanced funnel-aware pretraining, alternative pooling strategies, and comparisons with other model compression techniques to further optimize LLM efficiency.

\subsection{Future Work}

To address the identified limitations and deepen our understanding of funneling, several avenues of future research are promising. Expanding the investigation to include diverse model families beyond Gemma2 could enhance the generalizability of our conclusions. Further exploration into the effects of multiple funnel-aware pretraining layers would provide valuable insights into optimizing funnel configurations. Examining architectures that employ mixture-of-experts could also offer unique perspectives on managing information bottlenecks in complex models. Additionally, exploring alternative pooling operations could identify strategies that mitigate information loss more effectively. Lastly, directly comparing funneling with other established pruning or information bottleneck methods would clarify its relative strengths and weaknesses, guiding more effective deployment in various applications.

\subsection{Ethical Considerations}

The focus on efficiency should not overshadow broader ethical concerns inherent in LLM development, such as bias amplification, misuse potential, and the environmental impact of training large models, even if inference is optimized. To that end, we argue that more compact and efficient models are a step forward for inference-level environmental concerns, and reduce the chance of unintentional detail recitation. We have taken care not to use personally identifiable information in any of our training corpuses.

\newpage
% \bibliography{custom}
% \bibliographystyle{acl_natbib}

\appendix

\section{Appendix}
\label{sec:appendix}

Please see Tables \ref{table:hyperparameters} and \ref{tab:hyperparameters_7b} for Gemma2 2B and 7B hyperparameters, respectively. 

\begin{table*}[htb]
\centering
\caption{Hyperparameters for Gemma2 2B Model Training}
\label{table:hyperparameters}
\begin{tabular}{@{}ll@{}}
\toprule
Hyperparameter                      & Value                                  \\ \midrule
Gemma Model                        & GEMMA\_2B                               \\
Use Pooler                          & true                                   \\
Pooling Type                        & ATTENTION\_POOLING                     \\
Number of Attention Pooling Heads   & 4                                      \\
Number of Transformer Pooling Layers & 1                                      \\
Projection Dimension                  & None                                   \\
Max Steps                           & 2000                                   \\
Max Learning Rate                   & 0.00003                                \\
Minimum Learning Rate Fraction      & 0.1                                    \\
Warmup Steps                        & 100                                    \\
Funnel Pooling Config               & (1,1,1,1,1,1,2)                        \\
Sequence Length                     & 512                                    \\
Training Batch Size                 & 128                                    \\
Eval Batch Size                     & 256                                    \\
Number of Microbatches              & 0                                      \\
Weight Decay Rate                   & None (Likely 3.0 for Gemma)            \\
Soft Labels                         & False                                  \\
Scoring BF16 Mode                   & True                                   \\
Overtrain Multiplier                & 1                                      \\ \bottomrule
\end{tabular}
\end{table*}

\begin{table*}[htb]
\centering
\caption{Hyperparameters for Gemma2 7B Model Training}
\label{tab:hyperparameters_7b}
\begin{tabular}{@{}ll@{}}
\toprule
Hyperparameter                      & Value                                          \\ \midrule
Gemma Model                        & GEMMA\_7B                                       \\
Use Pooler                          & true                                           \\
Pooling Type                        & ATTENTION\_POOLING                             \\
Number of Attention Pooling Heads   & 4                                              \\
Number of Transformer Pooling Layers & 1                                              \\
Projection Dimension                  & None                                           \\
Max Steps                           & 4000                                           \\
Max Learning Rate                   & 0.00001                                        \\
Minimum Learning Rate Fraction      & 0.1                                            \\
Warmup Steps                        & 100                                            \\
Funnel Pooling Config               & (1,1,1,1,1,1,1,1,1,1,1,1,1,1,1,1,1,1,1,1,1,1,1,1,2) \\
Sequence Length                     & 512                                            \\
Training Batch Size                 & 128                                            \\
Eval Batch Size                     & 256                                            \\
Number of Microbatches              & 0                                              \\
Weight Decay Rate                   & None (Likely 3.0 for Gemma)                    \\
Soft Labels                         & False                                          \\
Scoring BF16 Mode                   & True                                           \\
Overtrain Multiplier                & 1                                              \\ \bottomrule
\end{tabular}
\end{table*}

\end{document}